\documentclass[twoside,11pt]{article}

\usepackage{blindtext}

\usepackage[preprint]{jmlr2e}

\usepackage{listings}
\usepackage[table,dvipsnames]{xcolor}
\usepackage{caption}
\usepackage{float}
\usepackage{booktabs}

\usepackage{textcomp}

\usepackage[most]{tcolorbox}
\usepackage{xcolor}

\definecolor{milblue}{RGB}{220,230,250}

\newcommand{\torchmil}{\texttt{torchmil}}
\newcommand{\torchmildata}{\texttt{torchmil.data}}
\newcommand{\torchmildatasets}{\texttt{torchmil.datasets}}
\newcommand{\torchmilmodels}{\texttt{torchmil.models}}
\newcommand{\torchmilnn}{\texttt{torchmil.nn}}

\newcommand*{\abmil}{ABMIL\@}
\newcommand*{\clam}{CLAM\@}
\newcommand*{\dsmil}{DSMIL\@}
\newcommand*{\patchgcn}{PatchGCN\@}
\newcommand*{\deepgraphsurv}{DeepGraphSurv\@}

\newcommand*{\transmil}{TransMIL\@}
\newcommand*{\dtfdmil}{DTFDMIL\@}
\newcommand*{\gtp}{GTP\@}
\newcommand*{\setmil}{SETMIL\@}
\newcommand*{\iibmil}{IIBMIL\@}

\newcommand*{\camil}{CAMIL\@}

\newcommand*{\smabmil}{SmABMIL\@}
\newcommand{\transformerabmil}{TransformerABMIL\@}
\newcommand{\smtransformerabmil}{SmTransformerABMIL\@}

\definecolor{codebg}{rgb}{0.99,0.99,0.99}
\definecolor{codeframe}{rgb}{0.8,0.8,0.8}
\definecolor{lightred}{HTML}{F5D2D2}
\definecolor{lightgreen}{HTML}{D2F5D2}

\lstdefinestyle{pythonstyle}{
    language=Python,
    backgroundcolor=\color{codebg},
    frame=single,
    rulecolor=\color{codeframe},
    basicstyle=\ttfamily\tiny,
    keywordstyle=\color{blue},
    commentstyle=\color{gray},
    stringstyle=\color{green!50!black},
    showstringspaces=false,
    breaklines=true
}

\usepackage{lastpage}

\ShortHeadings{torchmil: A PyTorch-based library for deep Multiple Instance Learning}{Castro-Macías and Sáez-Maldonado and Morales-Álvarez and Molina}
\firstpageno{1}

\begin{document}

\title{torchmil: A PyTorch-based library for deep Multiple Instance Learning}

\author{%
\name Francisco M Castro-Macías\textsuperscript{*\textbf{a}\textbf{b}} \email fcastro@ugr.es 
\vspace{1pt}\newline
\name Francisco J Sáez-Maldonado\textsuperscript{\textbf{a}\textbf{b}} \email fjaviersaezm@ugr.es
\vspace{1pt}\newline
\name Pablo Morales-Álvarez\textsuperscript{\textbf{c}\textbf{d}} \email pablomorales@ugr.es
\vspace{1pt}\newline
\name Rafael Molina\textsuperscript{\textbf{a}} \email rms@decsai.ugr.es
\vspace{1pt}\newline
{%
\textnormal{\textbf{*}}\textit{Corresponding author}\\
\textnormal{\textbf{\textsuperscript{\textbf{a}}}}\textit{Department of Computer Science and Artificial Intelligence, University of Granada, Spain}\\
\textnormal{\textbf{\textsuperscript{\textbf{b}}}}\textit{Research Centre for Information and Communications Technologies, University of Granada, Spain}\\
\textnormal{\textbf{\textsuperscript{\textbf{c}}}}\textit{Department of Statistics and Operations Research, University of Granada, Spain}\\
\textnormal{\textbf{\textsuperscript{\textbf{d}}}}\textit{Institute of Mathematics (IMAG), University of Granada, Spain}
}
}

\editor{My editor}

\maketitle

\begin{abstract}
Multiple Instance Learning (MIL) is a powerful framework for weakly supervised learning, particularly useful when fine-grained annotations are unavailable. 
Despite growing interest in deep MIL methods, the field lacks standardized tools for model development, evaluation, and comparison, which hinders reproducibility and accessibility. 
To address this, we present \torchmil, an open-source Python library built on PyTorch.
\torchmil\ offers a unified, modular, and extensible framework, featuring basic building blocks for MIL models, a standardized data format, and a curated collection of benchmark datasets and models. The library includes comprehensive documentation and tutorials to support both practitioners and researchers. 
\torchmil\ aims to accelerate progress in MIL and lower the entry barrier for new users.
Available at \textcolor{blue}{\url{https://torchmil.readthedocs.io}}.
\end{abstract}


\section{Introduction}

Multiple Instance Learning (MIL)~\citep{maron1997framework,gadermayr2024multiple} is a type of weakly supervised approach that is particularly helpful when fine-grained annotations are scarce. 
In MIL, training data is organized into labeled bags, each comprising multiple instances.
Unlike traditional supervised learning, which requires a label for every instance, in MIL labels are assigned only to bags, leaving instance labels unknown. 
In recent years, MIL has emerged as a highly active research area, with numerous contributions published in top-tier conferences and journals~\citep{zhang2022dtfd,fourkioti2023camil,castro2024sm,dure2025thinking}. 
Applications can be found in a broad range of areas, including computational pathology~\citep{song2023artificial,gadermayr2024multiple}, drug repositioning~\citep{gu2025deep}, and video event detection~\citep{lv2023unbiased}.

In recent years, a wide array of deep learning approaches has been proposed to tackle MIL problems. These span a variety of architectural paradigms, including transformer-based models~\citep{shao2021transmil}, graph neural networks~\citep{chen2021whole}, or the combination of both~\citep{castro2024sm}.
The complexity of MIL data makes the performance of these methods heavily dependent on preprocessing strategies and implementation details. 
Unfortunately, the fragmented and inconsistent nature of existing MIL codebases poses challenges for both reproducibility and accessibility, especially for newcomers to the field.

To address these challenges, we introduce \torchmil, an open-source Python library for deep MIL, built on top of PyTorch~\citep{paszke2017automatic}.
\torchmil\ provides a unified, modular, and extensible framework for building, training, and evaluating MIL models. 
It includes a set of reusable PyTorch modules tailored for MIL, a standardized representation for MIL data, and a growing collection of benchmark datasets and models.
In addition, \torchmil\ features tutorial notebooks and comprehensive documentation to support both beginners and advanced users.

Importantly, \torchmil\ is, to the best of our knowledge, the only existing framework that brings together both MIL datasets and models in a single environment.
Its aim is to serve as an accessible entry point for practitioners applying MIL to new domains, as well as a solid framework for researchers developing novel MIL methods. 
In this paper, we present the design principles and core features of \torchmil, along with a comprehensive empirical evaluation of the models currently implemented in the library.




\section{Library design and features}

In this section, we explain how \torchmil\ is designed, highlighting its main features.
\newline

\noindent
\textbf{Handling MIL data.}
Bags often differ in the number of instances, instance-level labels may be partially or entirely unavailable, and spatial or topological relationships among instances may need to be represented. 
Any data representation must support this information and allow efficient batching and parallel processing to enable scalable training.
To address these requirements, the \torchmildata\ submodule defines a standardized representation for MIL bags within \torchmil. Each bag is stored as a \texttt{TensorDict} object~\citep{bou2023torchrl}, in which each field encodes a specific property of the bag, such as instance features, the bag-level label, or a graph-based adjacency matrix capturing structural relationships. 
Batching is handled during the collation stage via an efficient padding and masking mechanism.
\newline

\noindent
\textbf{Datasets.}
When MIL data is not stored in a structured format -- for example, if instance-level information is fragmented -- data loading can become a computational bottleneck. 
To mitigate this, the \torchmildatasets\ submodule provides a recommended storage format and the \texttt{ProcessedMILDataset} class for efficient data access.
Moreover, at the time of writing, we have released three widely used MIL benchmark datasets on Hugging Face Datasets\footnote{\url{https://huggingface.co/torchmil}}: the RSNA Intracranial Hemorrhage Detection dataset~\citep{flanders2020construction}, the PANDA dataset~\citep{bulten2022artificial}, and the CAMELYON16 dataset~\citep{bejnordi2017diagnostic}.
Additionally, we include the algorithmic unit test datasets proposed by~\citet{raff2023reproducibility}.
We expect that this list of datasets will continue to grow.
\autoref{fig:torchmil_code} illustrates how these datasets can be integrated into a training pipeline.\newline


\noindent
\textbf{Modules and Models.}
A variety of deep MIL methods have been proposed in recent years, incorporating different mechanisms such as attention-based models (e.g., transformer-based architectures), graph neural networks, or combinations of both. 
In the \torchmilnn\ submodule, we provide modular PyTorch implementations of these core components, serving as foundational building blocks for constructing deep MIL models. 
Furthermore, the \torchmilmodels\ submodule includes implementations of 14 distinct popular MIL models (see~\autoref{table:experiments}). 
Each model is implemented as a subclass of the \texttt{MILModel} base class, which defines a unified interface for MIL model development within \torchmil.
We expect that this list of models will continue to grow.
In \autoref{fig:torchmil_code} we show how to instantiate one of these models. 
\newline


\noindent
\textbf{Documentation, examples, and tutorials.}
To support newcomers and promote a broader understanding of MIL, we have designed \torchmil\ to be accessible and easy to adopt. To this end, both the official repository and the project webpage provide a growing collection of examples and tutorials. These resources cover the fundamentals of \torchmil\ and illustrate how it can be applied across a variety of use cases.

\begin{figure}[H]
\centering
\begin{lstlisting}[style=pythonstyle]
import torch
from torchmil.datasets import Camelyon16MIL
from torchmil.models import TransformerABMIL
from torchmil.utils import Trainer
from torchmil.data import collate_fn
from torch.utils.data import DataLoader

# Load the Camelyon16 dataset
dataset = Camelyon16MIL(root="data", features="UNI")
dataloader = DataLoader(dataset, batch_size=4, shuffle=True, collate_fn=collate_fn)

# Instantiate the TransformerABMIL model and optimizer
model = TransformerABMIL(in_shape=(dataset.data_dim,), criterion=torch.nn.BCEWithLogitsLoss())
optimizer = torch.optim.Adam(model.parameters(), lr=1e-4)

# Instantiate the Trainer
trainer = Trainer(model, optimizer, device="cuda")

# Train the model
trainer.train(dataloader, epochs=10)

# Save the model
torch.save(model.state_dict(), "model.pt")
\end{lstlisting}
\vspace{-4mm}
\caption{
Example of training the TransformerABMIL model proposed by~\cite{castro2024sm} on the CAMELYON16 dataset.
\torchmil\ simplifies the process by providing built-in support for data preprocessing, loading, model configuration and training -- substantially reducing the amount of boilerplate code users need to write. 
}
\label{fig:torchmil_code}
\end{figure}

\section{Experiments}

In this section, we evaluate the quality of the implementations provided in \torchmil. 
We focus on the widely used CAMELYON16 benchmark dataset~\citep{bejnordi2017diagnostic}, which involves detecting breast cancer metastases from whole-slide images (WSIs).
We evaluate 14 deep MIL methods by comparing the \torchmil\ implementations against the original implementations released by their respective authors. The details about the experimental setup can be found in~\autoref{app:experiments}.

We include the following methods: 
\abmil~\citep{ilse2018attention}, \deepgraphsurv~\citep{li2018graph}, \clam ~\citep{lu2021data}, \dsmil ~\citep{li2021dual}, \patchgcn~\citep{chen2021whole}, \transmil ~\citep{shao2021transmil}, \dtfdmil ~\citep{zhang2022dtfd}, \setmil ~\citep{zhao2022setmil}, \gtp ~\citep{zheng2022graph}, \iibmil~\citep{ren2023iib}, \camil ~\citep{fourkioti2023camil}, \transformerabmil~(TABMIL, \cite{castro2024sm}), \smabmil~\citep{castro2024sm}, and \smtransformerabmil~(SmTABMIL, \cite{castro2024sm}).

The results are presented in \autoref{table:experiments}. 
For IIBMIL, we were unable to report results from the original implementation, as the authors did not release their training code. 
Our \torchmil\ implementation generally achieves comparable performance to the original implementations of the evaluated methods.
Notably, we observe two exceptions -- SETMIL and GTP -- where our implementation outperforms the original.
However, it is important to highlight that these methods were not originally evaluated on the CAMELYON16 dataset, which presents unique challenges and may require specific tuning.
In the case of SETMIL, the original implementation processes WSIs by cropping them to retain only a central region.
Given the large size of CAMELYON16 slides, this strategy risks excluding diagnostically relevant areas, which may explain the lower performance.
For GTP, the original implementation was tailored to relatively smaller WSIs, and its default hyperparameters may not generalize well to CAMELYON16.


\begin{figure}[t]
    \vspace{-5mm}
    \centering
    \includegraphics[width=0.8\textwidth]{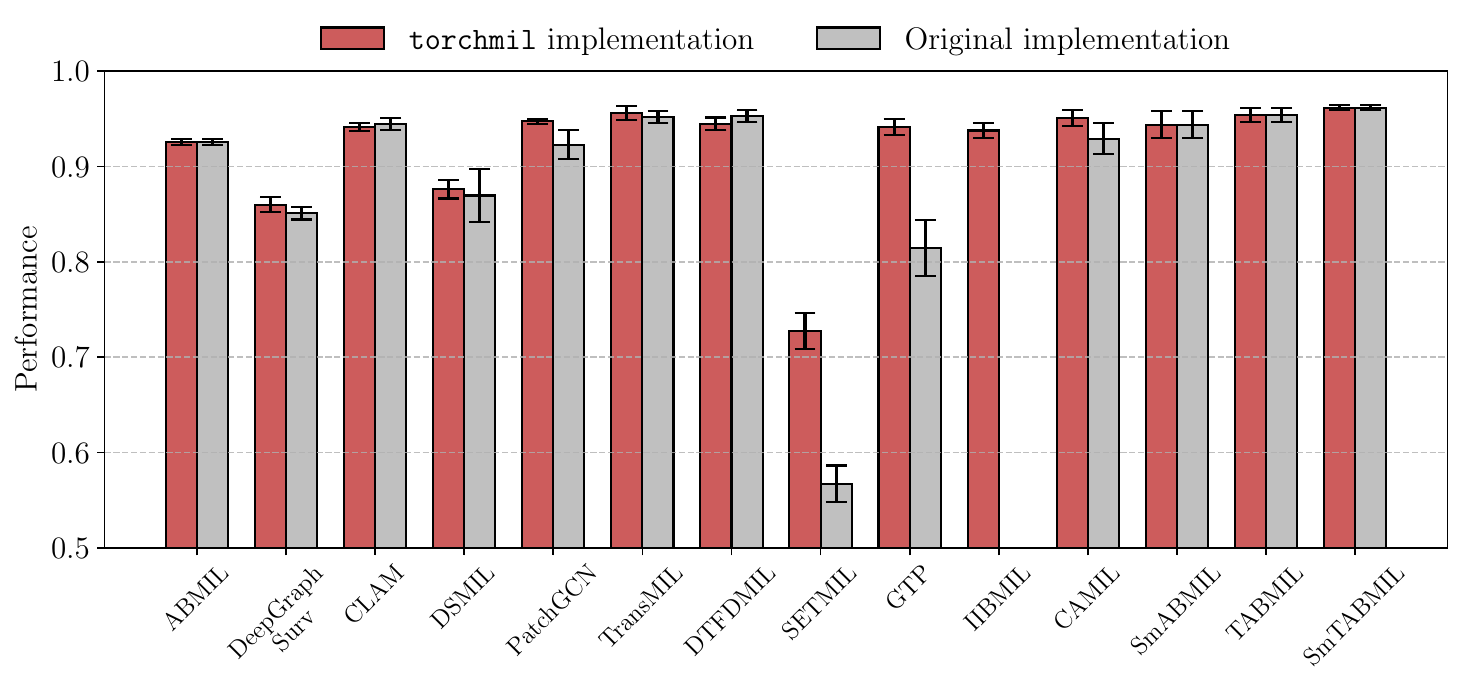}
    \vspace{-2mm}
    \caption{
    Performance comparison between the \torchmil\ implementation and the original implementations of various MIL models.
    Performance is reported as the average of accuracy, F1 score, and AUROC.
    The \torchmil\ implementation matches the original across all methods, except for SETMIL and GTP, where it performs better.
    See~\autoref{table:experiments} for the complete results.
    }
    \vspace{-3mm}
    \label{fig:comparison}
\end{figure}

\section{Conclusion}

In this work we introduced \torchmil, an open-source Python library for deep MIL.
Built on top of PyTorch, \torchmil\ provides a flexible and extensible framework for building, training, and evaluating deep MIL models.
It features a growing collection of core MIL components, datasets, and baseline models, designed to accelerate research and development in this area.
To support both newcomers and experienced practitioners, the library includes thorough documentation, practical tutorials, and example workflows.
With contributions from the community, we hope \torchmil\ becomes a collaborative platform to host new datasets and models, promote reproducibility and accessibility, and inspire future research in MIL.

\newpage

\acks{%
This work was supported by project PID2022-140189OB-C22 funded by MCIN / AEI / 10.13039 / 501100011033. 
Francisco M. Castro-Macías acknowledges FPU contract FPU21/01874 funded by Ministerio de Universidades. 
Pablo Morales-Álvarez acknowledges grant C-EXP-153-UGR23 funded by Consejería de Universidad, Investigación e Innovación and by the European Union (EU) ERDF Andalusia Program 2021-2027.
}


\vskip 0.2in
\bibliography{references}

\newpage

\appendix
\section{Experimental setup}
\label{app:experiments}

\textbf{Data preprocessing.}
Following \citet{lu2021data}, we extract patches of size $512 \times 512$ at $20\times$ magnification from each WSI. 
For each patch, features are obtained using a ResNet50 model pre-trained with the Barlow Twins self-supervised learning method~\citep{kang2023benchmarking}.
\newline

\noindent
\textbf{Training details.}
All models are trained using the same five train-validation splits and evaluated on the official CAMELYON16 test set.
Each method uses the original default hyperparameters as specified by its authors.
For our \torchmil\ implementations, training is performed within a unified framework: batch size of 1, Adam optimizer with a learning rate of $10^{-4}$, and 50 training epochs.
For the original implementations, we made only minimal modifications -- primarily to enable data loading -- while preserving each method’s original training code.

\begin{table}[h]
\centering
\resizebox{0.9\textwidth}{!}{%
\begin{tabular}{@{}c|ccc|ccc@{}}
\toprule
 & \multicolumn{3}{c|}{\torchmil\ implementation} & \multicolumn{3}{c}{Original implementation} \\
Model & ACC & AUROC & F1 & ACC & AUROC & F1 \\
\midrule
ABMIL & $0.922_{0.008}$ & $0.957_{0.003}$ & $0.896_{0.007}$ & $0.922_{0.008}$ & $0.957_{0.003}$ & $0.896_{0.007}$ \\
DeepGraphSurv & $0.864_{0.010}$ & $0.910_{0.011}$ & $0.805_{0.025}$ & $0.849_{0.008}$ & $0.898_{0.012}$ & $0.806_{0.020}$ \\
CLAM & $0.938_{0.008}$ & $0.969_{0.008}$ & $0.915_{0.010}$ & $0.939_{0.009}$ & $0.969_{0.015}$ & $0.924_{0.013}$ \\
DSMIL & $0.879_{0.015}$ & $0.928_{0.020}$ & $0.821_{0.023}$ & $0.888_{0.031}$ & $0.907_{0.022}$ & $0.813_{0.111}$ \\
PatchGCN & $0.947_{0.003}$ & $0.968_{0.007}$ & $0.926_{0.005}$ & $0.917_{0.028}$ & $0.967_{0.025}$ & $0.884_{0.040}$ \\
TransMIL & $0.954_{0.015}$ & $0.977_{0.007}$ & $0.938_{0.021}$ & $0.950_{0.012}$ & $0.973_{0.008}$ & $0.933_{0.017}$ \\
DTFDMIL & $0.939_{0.011}$ & $0.976_{0.014}$ & $0.918_{0.015}$ & $0.943_{0.017}$ & $0.979_{0.008}$ & $0.936_{0.013}$ \\
SETMIL & $0.775_{0.027}$ & $0.756_{0.042}$ & $0.652_{0.043}$ & $0.571_{0.048}$ & $0.559_{0.023}$ & $0.572_{0.044}$ \\
GTP & $0.940_{0.017}$ & $0.970_{0.010}$ & $0.915_{0.023}$ & $0.842_{0.045}$ & $0.821_{0.055}$ & $0.780_{0.079}$ \\
IIBMIL & $0.938_{0.018}$ & $0.960_{0.009}$ & $0.914_{0.022}$ & $\times$ & $\times$ & $\times$ \\
CAMIL & $0.948_{0.018}$ & $0.974_{0.008}$ & $0.929_{0.025}$ & $0.926_{0.036}$ & $0.961_{0.015}$ & $0.900_{0.045}$ \\
SmABMIL & $0.942_{0.037}$ & $0.971_{0.014}$ & $0.918_{0.036}$ & $0.942_{0.037}$ & $0.971_{0.014}$ & $0.918_{0.036}$ \\
TransformerABMIL & $0.952_{0.013}$ & $0.976_{0.013}$ & $0.934_{0.018}$ & $0.952_{0.013}$ & $0.976_{0.013}$ & $0.934_{0.018}$ \\
SmTransformerABMIL & $0.958_{0.004}$ & $0.982_{0.006}$ & $0.944_{0.006}$ & $0.958_{0.004}$ & $0.982_{0.006}$ & $0.944_{0.006}$ \\
\bottomrule
\end{tabular}
}
\caption{
Performance comparison between the \torchmil\ implementation and the original implementations of various MIL models.
A $\times$ symbol indicates that results are unavailable due to the original authors not releasing training code.
Except for SETMIL and GTP, \torchmil\ consistently matches the performance of the original implementation across every method.
}
\label{table:experiments}
\end{table}

\end{document}